\documentclass{sig-alt-full}
\def\tab{\ \ \ }

\begin{document}
%
\conferenceinfo{GECCO'09,} {July 8--12, 2009, Montr\'eal Qu\'ebec, Canada.} 
\CopyrightYear{2009}
\crdata{978-1-60558-325-9/09/07} 

\title{Efficient Natural Evolution Strategies}

\subtitle{Evolution Strategies and Evolutionary Programming Track}

%
%
%
%
%

\numberofauthors{4} 
%
\author{
%
%
\alignauthor Yi Sun\\
       \affaddr{IDSIA}\\
       \affaddr{Manno 6928, Switzerland}\\
       \email{yi@idsia.ch}
\alignauthor Daan Wierstra\\
       \affaddr{IDSIA}\\
       \affaddr{Manno 6928, Switzerland}\\
       \email{daan@idsia.ch}
\and  
\alignauthor Tom Schaul\\
       \affaddr{IDSIA}\\
       \affaddr{Manno 6928, Switzerland}\\
       \email{tom@idsia.ch}
\alignauthor J\"{u}rgen Schmidhuber\\
       \affaddr{IDSIA}\\
       \affaddr{Manno 6928, Switzerland}\\
       \email{juergen@idsia.ch}
}

\date{28 January 2009}

\maketitle
\begin{abstract}

Efficient Natural Evolution Strategies (eNES) is a novel alternative to 
conventional evolutionary algorithms, using the natural gradient
to adapt the mutation distribution. 
Unlike previous methods based on natural
gradients, eNES uses a  \emph{fast} algorithm to calculate 
the inverse of the \emph{exact} Fisher information matrix, thus increasing
both robustness and performance of its evolution gradient estimation,
even in higher dimensions. 
Additional novel aspects of eNES include optimal fitness baselines and importance mixing
(a procedure for updating the population with very few fitness evaluations). 
The algorithm yields competitive results on both unimodal and multimodal benchmarks.

\end{abstract}


\section{Introduction}
Evolutionary algorithms aim to optimize a `fitness' function 
that is either unknown or too complex to model directly.
They allow domain experts to search for good or near-optimal solutions to
numerous difficult real-world problems in areas ranging from medicine and
finance to control and robotics.
 
Typically, three objectives have to be kept in mind when developing
evolutionary algorithms---we want 
(1) robust performance; 
(2) few (potentially costly) fitness evaluations;
(3) scalability with problem dimensionality. 

We recently introduced Natural Evolution Strategies (NES; \cite{nespaper}), 
a new class of evolutionary algorithms less ad-hoc than traditional evolutionary methods. 
Here we propose a novel algorithm within this framework. It retains 
the theoretically well-founded nature of the original NES
while addressing its shortcomings w.r.t.~the above objectives.
 
NES algorithms maintain and iteratively update a multinormal mutation distribution. 
Parameters are updated by estimating a \emph{natural evolution gradient},
i.e.~the natural gradient on the parameters of the mutation distribution,
and following it towards better expected fitness.
Well-known advantages of natural gradient methods
include isotropic convergence on ill-shaped 
fitness landscapes~\cite{whynaturalamari}. This avoids drawbacks of
`vanilla' (regular) gradients which are prone to
 slow or premature convergence~\cite{nac}.

Our algorithm calculates the natural evolution gradient
using the \emph{exact} Fisher information matrix (FIM) and the Monte
Carlo-estimated gradient. 
In conjunction with the techniques of \emph{optimal fitness baselines} 
and \emph{fitness shaping} this yields robust performance (objective 1). 

To reduce the number of potentially costly evaluations (objective 2), we
introduce \emph{importance mixing}, a kind of steady-state enforcer which 
keeps the distribution of the new population conformed to the
current mutation distribution. 

To keep the computational cost manageable in
higher problem dimensions (objective 3), we derive a novel, efficient
algorithm for computing the inverse of the exact Fisher information matrix
(previous methods were either inefficient or approximate).
 
The resulting algorithm, \emph{Efficient Natural Evolution Strategies} (eNES), is elegant, requires no additional
heuristics and has few parameters that need tuning. 
It performs consistently well on both unimodal and multimodal benchmarks.

\section{Evolution Gradients}
First let us introduce the algorithm framework and the concept of evolution gradients.
The objective is to maximize a $d$-dimensional unknown fitness
function $f:\mathbb{R}^{d}\rightarrow \mathbb{R}$, while keeping the number
of function evaluations -- which are considered costly -- as low as
possible. The algorithm iteratively evaluates a population of size $n$ individuals $%
\mathbf{z}_{1}\ldots \mathbf{z}_{n}$ generated from the mutation distribution $%
p\left( \mathbf{z}|\theta \right) $. It then uses the fitness evaluations $f(%
\mathbf{z}_{1})\ldots f(\mathbf{z}_{n})$ to adjust parameters $\theta $ of
the mutation distribution.

Let $J\left( \theta \right) =\mathbb{E}\left[ f\left( \mathbf{z}\right)
|\theta \right] $ be the expected fitness under mutation distribution $p\left(
\mathbf{z}|\theta \right) $, namely,%
\begin{equation*}
J\left( \theta \right) =\int f\left( \mathbf{z}\right) p\left( \mathbf{z}%
|\theta \right) d\mathbf{z}\text{.}
\end{equation*}%
The core idea of our approach is to find, at each iteration, a small
adjustment $\delta \theta $, such that the expected fitness $J\left( \theta
+\delta \theta \right) $ is increased. The most straightforward approach is
to set $\delta \theta \propto \triangledown _{\theta }J\left( \theta \right)
$, where $\triangledown _{\theta }J\left( \theta \right) $ is the gradient
on $J\left( \theta \right) $. Using the `log likelihood trick', the gradient
can be written as%
\begin{align*}
\triangledown _{\theta }J\left( \theta \right) & =\triangledown _{\theta
}\int f\left( \mathbf{z}\right) p\left( \mathbf{z}|\theta \right) d\mathbf{z}
\\
& =\int f\left( \mathbf{z}\right) \triangledown _{\theta }p\left( \mathbf{z}%
|\theta \right) d\mathbf{z} \\
& =\int f\left( \mathbf{z}\right) \frac{p\left( \mathbf{z}|\theta \right) }{%
p\left( \mathbf{z}|\theta \right) }\triangledown _{\theta }p\left( \mathbf{z}%
|\theta \right) d\mathbf{z} \\
& =\int p\left( \mathbf{z}|\theta \right) \cdot \left( f\left( \mathbf{z}%
\right) \triangledown _{\theta }\ln p\left( \mathbf{z}|\theta \right)
\right) d\mathbf{z}\text{,}.
\end{align*}%
The last term can be approximated using Monte Carlo:%
\begin{equation*}
\triangledown _{\theta }^{s}J\left( \theta \right) =\frac{1}{n}%
\sum\nolimits_{i=1}^{n}f\left( \mathbf{z}_{i}\right) \triangledown _{\theta
}\ln p\left( \mathbf{z}_{i}|\theta \right) \text{,}
\end{equation*}%
where $\triangledown _{\theta }^{s}J\left( \theta \right) $ denotes the
estimated evolution gradient.

In our algorithm, we assume that $p\left( \mathbf{z}|\theta \right) $ is a
Gaussian distribution with parameters $\theta =\left\langle \mathbf{x},%
\mathbf{A}\right\rangle $, where $\mathbf{x}$ represents the mean, and $%
\mathbf{A}$ represents the Cholesky decomposition of the covariance matrix $%
\mathbf{C}$, such that $\mathbf{A}$ is upper triangular matrix and\footnote{%
For any matrix $\mathbf{Q}$, $\mathbf{Q}^{-}$ denotes its inverse and $%
\mathbf{Q}^{\top }$ denotes its transpose.} $\mathbf{C}=\mathbf{A}^{\top }%
\mathbf{A}$. The reason why we choose $\mathbf{A}$ instead of $\mathbf{C}$
as primary parameter is twofold. First, $\mathbf{A}$ makes explicit the $%
d\left( d+1\right) /2$ independent parameters determining the covariance
matrix $\mathbf{C}$. Second, the diagonal elements of $\mathbf{A}$ are the
square roots of the eigenvalues of $\mathbf{C}$, so $\mathbf{A}^{\top }%
\mathbf{A}$ is always positive semidefinite. In the rest of the text, we
assume $\theta $ is column vector of dimension $d_{s}=d+d\left( d+1\right)
/2 $ with elements in $\left\langle \mathbf{x},\mathbf{A}\right\rangle $
arranged as%
\begin{equation*}
\left[ \left( \theta ^{0}\right) ^{\top },\left( \theta ^{1}\right) ^{\top
}\ldots \left( \theta ^{d}\right) ^{\top }\right] ^{\top }\text{.}
\end{equation*}%
Here $\theta ^{0}=\mathbf{x}$ and $\theta ^{k}=\left[ a_{k,k}\ldots a_{k,d}%
\right] ^{\top }$ for $1\leq k\leq d$, where $a_{i,j}$ ($i\leq j$) denotes
the $\left( i,j\right) $-th element of $\mathbf{A}$.

Now we compute%
\begin{eqnarray*}
\mathbf{g}\left( \mathbf{z}|\theta \right)  &=&\triangledown _{\theta }\ln
p\left( \mathbf{z}|\theta \right)  \\
&=&\triangledown _{\theta }\{\frac{d}{2}\ln 2\pi -\frac{1}{2}\ln \left\vert 
\mathbf{A}\right\vert ^{2} \\
&&-\frac{1}{2}\left( \mathbf{A}^{-\top }\left( \mathbf{z}-\mathbf{x}\right)
\right) ^{\top }\left( \mathbf{A}^{-\top }\left( \mathbf{z}-\mathbf{x}%
\right) \right) \}\text{,}
\end{eqnarray*}
where $\mathbf{g}\left( \mathbf{z}|\theta \right) $ is assumed to be a $%
d_{s} $-dimensional column vector. The gradient w.r.t.~$\mathbf{x}$ is simply%
\begin{equation*}
\triangledown _{\mathbf{x}}\ln p\left( \mathbf{z}|\theta \right) =\mathbf{C}%
^{-}\left( \mathbf{z}-\mathbf{x}\right) \text{.}
\end{equation*}%
The gradient w.r.t.~$a_{i,j}$ ($i\leq j$) is given by%
\begin{equation*}
\frac{\partial }{\partial a_{i,j}}\ln p\left( \mathbf{z}|\theta \right)
=r_{i,j}-\delta \left( i,j\right) a_{i,i}^{-1}\text{,}
\end{equation*}%
where $r_{i,j}$ is the $\left( i,j\right) $-th element of%
\begin{equation*}
\mathbf{R}=\mathbf{A}^{-\top }\left( \mathbf{z}-\mathbf{x}\right) \left(
\mathbf{z-x}\right) ^{\top }\mathbf{C}^{-}
\end{equation*}%
and $\delta \left( i,j\right) $ is the Kronecker Delta function.

From $\mathbf{g}\left( \mathbf{z}|\theta \right) $, the mutation gradient $%
\triangledown _{\theta }^{s}J\left( \theta \right) $ can be computed as $%
\triangledown _{\theta }^{s}J\left( \theta \right) =\frac{1}{n}\mathbf{Gf}$,
where $\mathbf{G}=\left[ \mathbf{g}\left( \mathbf{z}_{1}|\theta \right)
\ldots \mathbf{g}\left( \mathbf{z}_{n}|\theta \right) \right] $, and $%
\mathbf{f}=\left[ f\left( \mathbf{z}_{1}\right) \ldots f\left( \mathbf{z}%
_{n}\right) \right] ^{\top }$. We update $\theta $ by $\delta \theta =\eta
\triangledown _{\theta }^{s}J\left( \theta \right) $, where $\eta $ is an
empirically tuned step size.

\section{Natural Gradient}
Vanilla gradient methods have been shown to converge slowly in fitness
landscapes with ridges and plateaus. Natural gradients~\cite{amari98natural}
constitute a principled approach for dealing with such problems. 
The natural gradient, unlike the vanilla
gradient, has the advantage of always pointing in the direction of the
steepest ascent. Furthermore, since the natural gradient is invariant w.r.t.
the particular parameterization of the mutation distribution, it
can cope with ill-shaped fitness landscapes and provides isotropic
convergence properties, which prevents premature convergence on plateaus and
avoids overaggressive steps on ridges~\cite{amari98natural}.

In this paper, we consider a special case of the natural gradient $\tilde{%
\triangledown}_{\theta }J$, defined as%
\begin{gather*}
\delta \theta ^{\top }\tilde{\triangledown}_{\theta }J=\underset{\delta \theta}{\max } J\left( \theta
+\delta \theta \right) \text{,} \\
\text{s.t.}\ KL\left( \theta +\delta \theta ||\theta \right) =\varepsilon
\text{,}
\end{gather*}%
where $\varepsilon $ is an arbitrarily small constant and $KL\left( \theta
^{\prime }||\theta \right) $ denotes the Kullback-Leibler divergence between
distributions $p\left( \mathbf{z}|\theta ^{\prime }\right) $ and $p\left(
\mathbf{z}|\theta \right) $. The constraints impose a geometry on $\theta $
which differs from the plain Euclidean one. With $\varepsilon \rightarrow 0$%
, the natural gradient $\tilde{\triangledown}_{\theta }J$ satisfies the
necessary condition $\mathbf{F}\tilde{\triangledown}_{\theta
}J=\triangledown _{\theta }J$, with $\mathbf{F}$ being the Fisher
information matrix:%
\begin{equation*}
\mathbf{F}=\mathbb{E}\left[ \triangledown _{\theta }\ln p\left( \mathbf{z}%
|\theta \right) \triangledown _{\theta }\ln p\left( \mathbf{z}|\theta
\right) ^{\top }\right] \text{.}
\end{equation*}

If $\mathbf{F}$ is invertible, which may not always be the case, the natural
gradient can be uniquely identified by $\tilde{\triangledown}_{\theta }J=%
\mathbf{F}^{-}\triangledown _{\theta }J$, or estimated from data using $%
\mathbf{F}^{-}\triangledown _{\theta }^{s}J$. The adjustment $\delta \theta $
can then be computed by%
\begin{equation*}
\delta \theta =\eta \mathbf{F}^{-}\triangledown _{\theta }^{s}J\text{.}
\end{equation*}

In the following sub-sections, we show that the FIM can in fact be computed
exactly, that it is invertible, and that there exists an efficient\footnote{%
Normally the FIM would involve $d_{s}^{2}=O\left( d^{4}\right) $ parameters,
which is intractable for most practical problems.} algorithm to compute the
inverse of the FIM.

\subsection{Derivation of the Exact FIM}

In the original NES \cite{nespaper}, we compute the natural evolution gradient
using the empirical Fisher information matrix, which is estimated
from the current population. This approach has three important
disadvantages. First, the empirical FIM is not guaranteed to be invertible,
which could result in unstable estimations. Second, a large population size
would be required to approximate the exact FIM up to a reasonable precision.
Third, it is highly inefficient to invert the empirical FIM, a matrix with $%
O\left( d^{4}\right) $ elements.

We circumvent these problems by computing the exact FIM directly from
mutation parameters $\theta $, avoiding the potentially unstable and
computationally costly method of estimating the empirical FIM from a
population which in turn was generated from $\theta $.

In eNES, the mutation distribution is the Gaussian defined by $\theta
=\left\langle \mathbf{x},\mathbf{A}\right\rangle $, the precise FIM $\mathbf{%
F}$ can be computed analytically. Namely, the $\left( m,n\right) $-th
element in $\mathbf{F}$ is given by%
\begin{equation*}
\left( \mathbf{F}\right) _{m,n}=\frac{\partial \mathbf{x}^{\top }}{\partial
\theta _{m}}\mathbf{C}^{-}\frac{\partial \mathbf{x}}{\partial \theta _{n}}+%
\frac{1}{2}\operatorname*{tr}\left( \mathbf{C}^{-}\frac{\partial \mathbf{C}}{%
\partial \theta _{m}}\mathbf{C}^{-}\frac{\partial \mathbf{C}}{\partial
\theta _{n}}\right) \text{,}
\end{equation*}%
where $\theta _{m}$, $\theta _{n}$ denotes the $m$-th and $n$-th element in $%
\theta $. Let $i_{m},j_{m}$ be the $a_{i_{m},j_{m}}$ such that it appears at
the $\left( d+m\right) $-th position in $\theta $. First, notice that%
\begin{equation*}
\frac{\partial \mathbf{x}^{\top }}{\partial x_{i}}\mathbf{C}^{-}\frac{%
\partial \mathbf{x}}{\partial x_{j}}=\left( \mathbf{C}^{-}\right) _{i,j}%
\text{,}
\end{equation*}%
and%
\begin{equation*}
\frac{\partial \mathbf{x}^{\top }}{\partial a_{i_{1},j_{1}}}\mathbf{C}^{-}%
\frac{\partial \mathbf{x}}{\partial a_{i_{2},j_{2}}}=\frac{\partial \mathbf{x%
}^{\top }}{\partial x_{i}}\mathbf{C}^{-}\frac{\partial \mathbf{x}}{\partial
a_{j,k}}=0\text{.}
\end{equation*}%
So the upper left corner of the FIM is $\mathbf{C}^{-}$, and $\mathbf{F}$
has the following shape%
\begin{equation*}
\mathbf{F}=\left[ 
\begin{array}{cc}
\mathbf{C}^{-} & \mathbf{0} \\ 
\mathbf{0} & \mathbf{F}_{\mathbf{A}}%
\end{array}%
\right] \text{.}
\end{equation*}%
The next step is to compute $\mathbf{F}_{\mathbf{A}}$. Note that%
\begin{equation*}
\left( \mathbf{F}_{\mathbf{A}}\right) _{m,n}=\frac{1}{2}\operatorname*{tr}\left[ 
\mathbf{C}^{-}\frac{\partial \mathbf{C}}{\partial a_{i_{m},j_{m}}}\mathbf{C}%
^{-}\frac{\partial \mathbf{C}}{\partial a_{i_{n},j_{n}}}\right] \text{.}
\end{equation*}%
Using the relation%
\begin{equation*}
\frac{\partial \mathbf{C}}{\partial a_{i,j}}=\frac{\partial }{\partial
a_{i,j}}\mathbf{A}^{\top }\mathbf{A=}\frac{\partial \mathbf{A}^{\top }}{%
\partial a_{i,j}}\mathbf{A+A}^{\top }\frac{\partial \mathbf{A}}{\partial
a_{i,j}}\text{,}
\end{equation*}%
and the properties of the trace, we get

\begin{eqnarray*}
\left( \mathbf{F}_{\mathbf{A}}\right) _{m,n} &=&\operatorname*{tr}\left[ \mathbf{A}%
^{-}\frac{\partial \mathbf{A}}{\partial a_{i_{m},j_{m}}}\mathbf{A}^{-}\frac{%
\partial \mathbf{A}}{\partial a_{i_{n},j_{n}}}\right]  \\
&&+\operatorname*{tr}\left[ \frac{\partial \mathbf{A}}{\partial a_{i_{m},j_{m}}}%
\mathbf{C}^{-}\frac{\partial \mathbf{A}^{\top }}{\partial a_{i_{n},j_{n}}}%
\right] \text{.}
\end{eqnarray*}%
Computing the first term gives us%
\begin{equation*}
\operatorname*{tr}\left[ \mathbf{A}^{-}\frac{\partial \mathbf{A}}{\partial
a_{i_{m},j_{m}}}\mathbf{A}^{-}\frac{\partial \mathbf{A}}{\partial
a_{i_{n},j_{n}}}\right] =\left( \mathbf{A}^{-}\right) _{j_{n},i_{m}}\left( 
\mathbf{A}^{-}\right) _{j_{m},i_{n}}\text{.}
\end{equation*}%
Note that since $\mathbf{A}$ is upper triangular, $\mathbf{A}^{-}$ is also
upper triangular, so the first summand is non-zero iff%
\begin{equation*}
i_{n}=i_{m}=j_{n}=j_{m}\text{.}
\end{equation*}%
In this case, $\left( \mathbf{A}^{-}\right) _{j_{n},i_{m}}=\left( \mathbf{A}%
^{-}\right) _{j_{m},i_{n}}=a_{j_{n},i_{m}}^{-1}$, so%
\begin{equation*}
\operatorname*{tr}\left[ \mathbf{A}^{-}\frac{\partial \mathbf{A}}{\partial
a_{i_{m},j_{m}}}\mathbf{A}^{-}\frac{\partial \mathbf{A}}{\partial
a_{i_{n},j_{n}}}\right] =a_{i_{m},i_{n}}^{-2}\delta \left(
i_{m},i_{n},j_{m},j_{n}\right) \text{.}
\end{equation*}%
Here $\delta \left( \cdot \right) $ is the generalized Kronecker Delta
function, i.e. $\delta \left( i_{m},i_{n},j_{m},j_{n}\right)=1$ iff all
four indices are the same. The second term is computed as%
\begin{equation*}
\operatorname*{tr}\left[ \frac{\partial \mathbf{A}}{\partial a_{i_{m},j_{m}}}%
\mathbf{C}^{-}\frac{\partial \mathbf{A}^{\top }}{\partial a_{i_{n},j_{n}}}%
\right] =\left( \mathbf{C}^{-}\right) _{j_{n},j_{m}}\delta \left(
i_{n},i_{m}\right) \text{.}
\end{equation*}%
Therefore, we have%
\begin{equation*}
\left( \mathbf{F}_{\mathbf{A}}\right) _{m,n}=\left( \mathbf{C}^{-}\right)
_{j_{n},j_{m}}\delta \left( i_{n},i_{m}\right) +a_{i_{m},i_{n}}^{-2}\delta
\left( i_{m},i_{n},j_{m},j_{n}\right) \text{.}
\end{equation*}%
It can easily be proven that $\mathbf{F}_{\mathbf{A}}$ itself is a block
diagonal matrix with $d$ blocks along the diagonal, with sizes ranging from $%
d$ to $1$. Therefore, the precise FIM is given by%
\begin{equation*}
\mathbf{F=}\left[ 
\begin{array}{cccc}
\mathbf{F}_{0} &  &  &  \\ 
& \mathbf{F}_{1} &  &  \\ 
&  & \ddots  &  \\ 
&  &  & \mathbf{F}_{d}%
\end{array}%
\right] \text{,}
\end{equation*}%
with $\mathbf{F}_{0}=\mathbf{C}^{-}$ and block $\mathbf{F}_{k}$ ($d\geq
k\geq 1$) given by%
\begin{equation*}
\mathbf{F}_{k}=\left[ 
\begin{array}{cc}
a_{k,k}^{-2} & \mathbf{0} \\ 
\mathbf{0} & \mathbf{0}%
\end{array}%
\right] +\mathbf{D}_{k}\text{.}
\end{equation*}%
Here $\mathbf{D}_{k}$ is the lower-right square submatrix of $\mathbf{C}^{-}$
with dimension $d+1-k$, e.g.~$\mathbf{D}_{1}=\mathbf{C}^{-}$, and $\mathbf{D%
}_{d}=\left( \mathbf{C}^{-}\right) _{d,d}$.

We prove that the FIM given above is invertible if $\mathbf{C}$ is
invertible. $\mathbf{F}_{k}$ $\left( 1\leq k\leq d\right) $ being invertible
follows from the fact that the submatrix $\mathbf{D}_{k}$ on the main
diagonal of a positive definite matrix $\mathbf{C}^{-}$ must also be
positive definite, and adding $a_{k,k}^{-2}>0$ to the diagonal would not
decrease any of its eigenvalues. Also note that $\mathbf{F}_{0}=\mathbf{C}%
^{-}$ is invertible, so $\mathbf{F}$ is invertible.

It is worth pointing out that the block diagonal structure of $\mathbf{F}$
partitions parameters $\theta $ into $d+1$ orthogonal groups $\theta
^{0}\ldots \theta ^{k}$, which suggests that we could modify each group of
parameters without affecting other groups. We will need this intuition in
the next section.

\subsection{Iterative Computation of FIM Inverse}

The exact FIM is a block diagonal matrix with $d+1$ blocks. Normally,
inverting the FIM requires $d$ matrix inversions. However, we can explore
the structure of each sub-block in order to make the inverse of $\mathbf{F}$
more efficient, both in terms of time and space complexity.

First, we realize that $\mathbf{F}_{d}$ is simply a number, so its inversion
is given by $\mathbf{F}_{d}^{-}=\left( \left( \mathbf{C}^{-}\right)
_{d,d}+a_{d,d}^{-2}\right) ^{-1}$, and similarly $\mathbf{D}_{d}^{-}=\left(
\left( \mathbf{C}^{-}\right) _{d,d}\right) ^{-1}$. Now, letting $k$ vary
from $d-1$ to $1$, we can compute $\mathbf{F}_{k}^{-}$ and $\mathbf{D}%
_{k}^{-}$ directly from $\mathbf{D}_{k+1}^{-}$. By block matrix inversion%
\begin{equation*}
\left[ 
\begin{array}{cc}
\mathbf{P}_{11} & \mathbf{P}_{12} \\ 
\mathbf{P}_{21} & \mathbf{P}_{22}%
\end{array}%
\right] ^{-}=\left[ 
\begin{array}{cc}
\mathbf{Q}_{1}^{-} & -\mathbf{P}_{11}^{-}\mathbf{P}_{12}\mathbf{Q}_{2}^{-}
\\ 
-\mathbf{Q}_{2}^{-}\mathbf{P}_{21}\mathbf{P}_{11}^{-} & \mathbf{Q}_{2}^{-}%
\end{array}%
\right] \text{,}
\end{equation*}%
using%
\begin{equation*}
\mathbf{Q}_{1}=\mathbf{P}_{11}-\mathbf{P}_{12}\mathbf{P}_{22}^{-}\mathbf{P}%
_{21}\text{, }\mathbf{Q}_{2}=\mathbf{P}_{22}-\mathbf{P}_{21}\mathbf{P}%
_{11}^{-}\mathbf{P}_{12}\text{,}
\end{equation*}%
and the Woodbury identity%
\begin{align*}
\mathbf{Q}_{2}^{-}& =\left[ \mathbf{P}_{22}+\mathbf{P}_{21}\left( -\mathbf{P}%
_{11}^{-}\right) \mathbf{P}_{21}^{\top }\right] ^{-} \\
& =\mathbf{P}_{22}^{-}-\mathbf{P}_{22}^{-}\mathbf{P}_{21}\left[ -\mathbf{P}%
_{11}+\mathbf{P}_{21}^{\top }\mathbf{P}_{22}^{-}\mathbf{P}_{21}\right] ^{-}%
\mathbf{P}_{21}^{\top }\mathbf{P}_{22}^{-}\text{,}
\end{align*}%
(also noting that in our case, $\mathbf{P}_{11}$ is a number $\left( \mathbf{%
C}^{-}\right) _{k,k}$), we can state 
\begin{equation*}
\mathbf{Q}_{2}^{-}=\mathbf{P}_{22}^{-}-\frac{\left( \mathbf{P}_{22}^{-}%
\mathbf{P}_{21}\right) \left( \mathbf{P}_{22}^{-}\mathbf{P}_{21}\right)
^{\top }}{\mathbf{P}_{21}^{\top }\mathbf{P}_{22}^{-}\mathbf{P}_{21}-\mathbf{P%
}_{11}}\text{.}
\end{equation*}%
This can be computed directly from $\mathbf{P}_{22}^{-}$, i.e.~$\mathbf{D}%
_{k+1}^{-}$. Skipping the intermediate steps, we propose the following
algorithm for computing $\mathbf{F}_{k}^{-}$ and $\mathbf{D}_{k}^{-}$ from $%
\mathbf{D}_{k+1}^{-}$:%
\begin{align*}
\mathbf{v}& =\left( \mathbf{C}^{-}\right) _{k+1:d,k}\text{, }w=\left( 
\mathbf{C}^{-}\right) _{k,k}\text{, }\mathbf{u}=\mathbf{D}_{k+1}^{-}\mathbf{v%
}\text{,} \\
s& =\mathbf{v}^{\top }\mathbf{u}\text{, }q=\left( w-s\right) ^{-1}\text{, }%
q_{F}=\left( w_{F}-s\right) ^{-1}\text{,} \\
c& =-w^{-1}\left( 1+qs\right) \text{, }c_{F}=-w_{F}^{-1}\left(
1+q_{F}s\right) \text{,} \\
\mathbf{F}_{k}^{-}& =\left[ 
\begin{array}{cc}
q_{F} & c_{F}\mathbf{u}^{\top } \\ 
c_{F}\mathbf{u}^{\top } & \mathbf{D}_{k+1}^{-}+q_{F}\mathbf{uu}^{\top }%
\end{array}%
\right] \text{,} \\
\mathbf{D}_{k}^{-}& =\left[ 
\begin{array}{cc}
q & c\mathbf{u}^{\top } \\ 
c\mathbf{u}^{\top } & \mathbf{D}_{k+1}^{-}+q\mathbf{uu}^{\top }%
\end{array}%
\right] \text{.}
\end{align*}%
Here $\left( \mathbf{C}^{-}\right) _{k+1:d,k}$ is the sub-vector in $\mathbf{%
C}^{-}$ at column $k$, and row $k+1$ to $d$. A single update from $\mathbf{D}%
_{k+1}^{-}$ to $\mathbf{F}_{k}^{-}$ and $\mathbf{D}_{k}^{-}$ requires $%
O\left( \left( d-k\right) ^{2}\right) $ floating point multiplications. The
overall complexity of computing all sub-blocks $\mathbf{F}_{k}^{-}$, $1\leq
k\leq d$, is thus $O\left( d^{3}\right) $.

The algorithm is efficient both in time and storage in the sense that, on
one hand, there is no explicit matrix inversion, while on the other hand,
the space for storing $\mathbf{D}_{k}$ (including $\mathbf{F}_{k}$, if not
needed later) can be reclaimed immediately after each iteration, which means
that at most $O\left( d^{2}\right) $ storage is required. Note also that $%
\mathbf{F}_{k}^{-}$ can be used directly to compute $\delta \theta ^{k}$,
using $\delta \theta ^{k}=\mathbf{F}_{k}^{-}\mathbf{G}^{k}\mathbf{f}$, where 
\begin{align*}
\mathbf{G}^{k}& =\left[ \mathbf{g}^{k}\left( \mathbf{z}_{1}\right) ,\ldots ,%
\mathbf{g}^{k}\left( \mathbf{z}_{n}\right) \right]  \\
& =\left[ \triangledown _{\theta ^{k}}\ln p\left( \mathbf{z}|\theta \right)
,\ldots ,\triangledown _{\theta ^{k}}\ln p\left( \mathbf{z}|\theta \right) %
\right] 
\end{align*}%
is the submatrix of $\mathbf{G}$ w.r.t. the mutation gradient of $\theta ^{k}
$.

To conclude, the algorithm given above efficiently computes the inverse of
the exact FIM, required for computing the natural mutation gradient.

\section{Optimal Fitness Baselines}
\label{baselines}
The concept of \emph{fitness baselines}, first introduced in~\cite{nespaper}, constitutes an efficient variance reduction method
for estimating $\delta \theta $. However, baselines as found in~\cite{petersthesis} 
are suboptimal w.r.t.~the variance of $\delta \theta $, since
this FIM\ may not be invertible. It is difficult to formulate the variance
of $\delta \theta $ directly. However, since the exact FIM is invertible and
can be computed efficiently, we can in fact compute an optimal baseline for
minimizing the variance of $\delta \theta $, given by%
\begin{eqnarray*}
\operatorname*{Var}\left( \delta \theta \right) &=&\eta ^{2}\mathbb{E[}\left( \mathbf{F}%
^{-}\triangledown _{\theta }^{s}J-\mathbb{E}\left[ \mathbf{F}%
^{-}\triangledown _{\theta }^{s}J\right] \right) ^{\top } \\
&&\cdot \left( \mathbf{F}^{-}\triangledown _{\theta }^{s}J-\mathbb{E}\left[
\mathbf{F}^{-}\triangledown _{\theta }^{s}J\right] \right) ]\text{,}
\end{eqnarray*}%
where $\triangledown _{\theta }^{s}J$ is the estimated evolution gradient,
which is given by%
\begin{equation*}
\triangledown _{\theta }^{s}J=\frac{1}{n}\sum\nolimits_{i=1}^{n}\left[
f\left( z_{i}\right) -b\right] \triangledown _{\theta }\ln p\left( \mathbf{z}%
_{i}|\theta \right) \text{.}
\end{equation*}%
The scalar $b$ is called the fitness baseline. Adding $b$ does not affect
the expectation of $\triangledown _{\theta }^{s}J$, since%
\begin{eqnarray*}
\mathbb{E}\left[ \triangledown _{\theta }^{s}J\right] &=&\triangledown
_{\theta }\int \left( f\left( \mathbf{z}\right) -b\right) p\left( \mathbf{z}%
|\theta \right) d\mathbf{z} \\
&=&\triangledown _{\theta }\int f\left( \mathbf{z}\right) p\left( \mathbf{z}%
|\theta \right) d\mathbf{z}\text{.}
\end{eqnarray*}%
However, the variance depends on the value of $b$, i.e.%
\begin{eqnarray*}
\operatorname*{Var}\left( \delta \theta \right) &\propto &b^{2}\mathbb{E}\left[ \left(
\mathbf{F}^{-}\mathbf{G1}\right) ^{\top }\left( \mathbf{F}^{-}\mathbf{G1}%
\right) \right] \\
&&-2b\mathbb{E}\left[ \left( \mathbf{F}^{-}\mathbf{Gf}\right) ^{\top }\left(
\mathbf{F}^{-}\mathbf{G1}\right) \right] +\text{const.}
\end{eqnarray*}%
Here $\mathbf{1}$ denotes a $n$-by-$1$ vector filled with $1$s. The optimal
value of the baseline is%
\begin{equation*}
b=\frac{\mathbb{E}\left[ \left( \mathbf{F}^{-}\mathbf{Gf}\right) ^{\top
}\left( \mathbf{F}^{-}\mathbf{G1}\right) \right] }{\mathbb{E}\left[ \left(
\mathbf{F}^{-}\mathbf{G1}\right) ^{\top }\left( \mathbf{F}^{-}\mathbf{G1}%
\right) \right] }\text{.}
\end{equation*}%
Assuming individuals are i.i.d., $b$ can be approximated from data by%
\begin{equation*}
b\simeq \frac{\sum_{i=1}^{n}f\left( \mathbf{z}_{i}\right) \left( \mathbf{F}%
^{-}\mathbf{g}\left( \mathbf{z}_{i}\right) \right) ^{\top }\left( \mathbf{F}%
^{-}\mathbf{g}\left( \mathbf{z}_{i}\right) \right) }{\sum_{i=1}^{n}\left(
\mathbf{F}^{-}\mathbf{g}\left( \mathbf{z}_{i}\right) \right) ^{\top }\left(
\mathbf{F}^{-}\mathbf{g}\left( \mathbf{z}_{i}\right) \right) }\text{.}
\end{equation*}

In order to further reduce the estimation covariance, we can utilize a
parameter-specific baseline for each parameter $\theta _{j}$ individually,
which is given by%
\begin{equation*}
b_{j}=\frac{\mathbb{E}\left[ \left( \mathbf{h}_{j}\mathbf{Gf}\right) \left(
\mathbf{h}_{j}\mathbf{G1}\right) \right] }{\mathbb{E}\left[ \left( \mathbf{h}%
_{j}\mathbf{G1}\right) \left( \mathbf{h}_{j}\mathbf{G1}\right) \right] }%
\simeq \frac{\sum_{i=1}^{n}f\left( \mathbf{z}_{i}\right) \left( \mathbf{h}%
_{j}\mathbf{g}\left( \mathbf{z}_{i}\right) \right) ^{2}}{\sum_{i=1}^{n}%
\left( \mathbf{h}_{j}\mathbf{g}\left( \mathbf{z}_{i}\right) \right) ^{2}}%
\text{.}
\end{equation*}%
Here $\mathbf{h}_{j}$ is the $j$-th row vector of $\mathbf{F}^{-}$.

However, parameter-specific baseline values $\theta _{j}$ might reduce
variance too much, which harms the performance of the algorithm.
Additionally, adopting different baseline values for correlated parameters
may affect the underlying structure of the parameter space, rendering
estimations unreliable. To address both of these problems, we follow the
intuition that if the $\left( m,n\right) $-th element in the FIM is $0$,
then parameters $\theta _{m}$ and $\theta _{n}$ are orthogonal and only
weakly correlated. Therefore, we propose using the \emph{block fitness
baseline}, i.e.~a single baseline $b^{k}$ for each group of parameters $%
\theta ^{k}$, $0\leq k\leq d$. Its formulation is given by%
\begin{align*}
b^{k}& =\frac{\mathbb{E}\left[ \left( \mathbf{F}_{k}^{-}\mathbf{G}^{k}%
\mathbf{f}\right) \left( \mathbf{F}_{k}^{-}\mathbf{G}^{k}\mathbf{1}\right) %
\right] }{\mathbb{E}\left[ \left( \mathbf{F}_{k}^{-}\mathbf{G}^{k}\mathbf{1}%
\right) \left( \mathbf{F}_{k}^{-}\mathbf{G}^{k}\mathbf{1}\right) \right] } \\
& \simeq \frac{\sum_{i=1}^{n}f\left( \mathbf{z}_{i}\right) \left( \mathbf{F}%
_{k}^{-}\mathbf{g}^{k}\left( \mathbf{z}_{i}\right) \right) ^{\top }\left(
\mathbf{F}_{k}^{-}\mathbf{g}^{k}\left( \mathbf{z}_{i}\right) \right) }{%
\sum_{i=1}^{n}\left( \mathbf{F}_{k}^{-}\mathbf{g}^{k}\left( \mathbf{z}%
_{i}\right) \right) ^{\top }\left( \mathbf{F}_{k}^{-}\mathbf{g}^{k}\left(
\mathbf{z}_{i}\right) \right) }\text{.}
\end{align*}%
Here $\mathbf{F}_{k}^{-}$ denotes the inverse of the $k$-th diagonal block
of $\mathbf{F}^{-}$, while $\mathbf{G}^{k}$ and $\mathbf{g}^{k}$ denote the
submatrices corresponding to differentiation w.r.t.~$\theta ^{k}$. 

In a companion paper~\cite{nespapericml}, we empirically investigated the
convergence properties when using the various types of baseline. We found
block fitness baselines to be very robust, whereas uniform and parameter-specific
baselines sometimes led to premature convergence.


The main complexity for computing the optimal fitness baseline pertains to
the necessity of storing a potentially large gradient matrix $\mathbf{G}$,
with dimension $d_{s}\times n\sim O\left( nd^{2}\right) $. The time
complexity, in this case, is $O\left( nd^{3}\right) $ since we have to
multiply each $\mathbf{F}_{k}^{-}$ with $\mathbf{G}^{k}$. For large problem
dimensions, the storage requirement may not be acceptable since $n$ also
scales with $d$. We solve this problem by introducing a time-space tradeoff
which reduces the storage requirement to $O\left( d^{2}\right) $ while
keeping the time complexity unchanged. In particular, we first compute for
each $k$, a scalar $u_{k}=\mathbf{a}_{k}^{-}\left( \mathbf{z}-\mathbf{x}%
\right) $, where $\mathbf{a}_{k}^{-}$ is the $k$-th row vector of $\mathbf{A}%
^{-}$. Then, for $1\leq i\leq n$, we compute the vector $\mathbf{v}=\left( 
\mathbf{C}^{-}\right) _{k:d}\left( \mathbf{z}-\mathbf{x}\right) $, where $%
\left( \mathbf{C}^{-}\right) _{k:d}$ is the submatrix of $\mathbf{C}^{-}$ by
taking rows $k$ to $d$. The gradient $\mathbf{g}^{k}\left( \mathbf{z}%
_{i}\right) $ can be computed from $u_{k}$ and $\mathbf{v}$, and used to
compute $\mathbf{F}_{k}^{-}\mathbf{g}^{k}\left( \mathbf{z}_{i}\right) $
directly. The storage for $\mathbf{g}^{k}\left( \mathbf{z}_{i}\right) $ can
be immediately reclaimed. Finally, the complexity of computing $\mathbf{g}%
^{k}\left( \mathbf{z}_{i}\right) $ for all $i$ is $O\left( nd\left(
d-k\right) \right) $, so the total complexity of computing every element in $%
\mathbf{G}$ would still be $O\left( nd^{3}\right) $.

\section{Importance mixing}
\label{importance}
At each generation, we evaluate $n$ new individuals generated from mutation
distribution $p\left( \mathbf{z}|\theta \right) $. However, since small
updates ensure that the KL divergence between consecutive mutation
distributions is generally small, most new individuals will fall in the high
density area of the previous mutation distribution $p\left( \mathbf{z}%
|\theta ^{\prime }\right) $. This leads to redundant fitness evaluations in
that same area.

Our solution to this problem is a new procedure called importance mixing,
which aims to \emph{reuse} fitness evaluations from the previous generation,
while ensuring the updated population conforms to the new mutation
distribution.

Importance mixing works in two steps: In the first step, rejection sampling
is performed on the previous population, such that individual $\mathbf{z}$
is accepted with probability%
\begin{equation*}
\min \left\{ 1,\left( 1-\alpha \right) \frac{p\left( \mathbf{z}|\theta
\right) }{p\left( \mathbf{z}|\theta ^{\prime }\right) }\right\} \text{.}
\end{equation*}%
Here $\alpha \in \left[ 0,1\right] $ is the \emph{minimal refresh rate}. Let 
$n_{a}$ be the number of individuals accepted in the first step. In the
second step, reverse rejection sampling is performed as follows: Generate
individuals from $p\left( \mathbf{z}|\theta \right) $ and accept $\mathbf{z}$
with probability%
\begin{equation*}
\max \left\{ \alpha ,1-\frac{p\left( \mathbf{z}|\theta ^{\prime }\right) }{%
p\left( \mathbf{z}|\theta \right) }\right\} 
\end{equation*}%
until $n-n_{a}$ new individuals are accepted. The $n_{a}$ individuals from
the old generation and $n-n_{a}$ newly accepted individuals together
constitute the new population. Note that only the fitnesses of the newly
accepted individuals need to be evaluated. The advantage of using importance
mixing is twofold: On the one hand, we reduce the number of fitness
evaluations required in each generation, on the other hand, if we fix the
number of newly evaluated fitnesses, then many more fitness
evaluations can potentially be used to yield more reliable and accurate gradients.

The minimal refresh rate $\alpha $ lower bounds the expected proportion of
newly evaluated individuals $\rho =\mathbb{E}\left[ \frac{n-n_{a}}{n}\right] 
$, namely $\rho \geq \alpha $, with the equality holding iff $\theta =\theta
^{\prime }$. In particular, if $\alpha =1$, all individuals from the
previous generation will be discarded, and if $\alpha =0$, $\rho $ depends
only on the distance between $p\left( \mathbf{z}|\theta \right) $ and $%
p\left( \mathbf{z}|\theta ^{\prime }\right) $. Normally we set $\alpha $ to
be a small positive number, e.g.~$0.01$, to avoid too low an acceptance
probability at the second step when $p\left( \mathbf{z}|\theta ^{\prime
}\right) /p\left( \mathbf{z}|\theta \right) \simeq 1$.

It can be proven that the updated population conforms to the mutation
distribution $p\left( \mathbf{z}|\theta \right) $. In the region where $%
\left( 1-\alpha \right) p\left( \mathbf{z}|\theta \right) /p\left( \mathbf{z}%
|\theta ^{\prime }\right) \leq 1$, the probability that an individual from
previous generations appears in the new population is%
\begin{equation*}
p\left( \mathbf{z}|\theta ^{\prime }\right) \cdot \left( 1-\alpha \right)
p\left( \mathbf{z}|\theta \right) /p\left( \mathbf{z}|\theta ^{\prime
}\right) =\left( 1-\alpha \right) p\left( \mathbf{z}|\theta \right) \text{.}
\end{equation*}%
The probability that an individual generated from the second step entering
the population is $\alpha p\left( \mathbf{z}|\theta \right) $, since%
\begin{equation*}
\max \left\{ \alpha ,1-p\left( \mathbf{z}|\theta ^{\prime }\right) /p\left( 
\mathbf{z}|\theta \right) \right\} =\alpha.
\end{equation*}
So the probability of an
individual entering the population is just $p\left( \mathbf{z}|\theta
\right) $ in that region. The same result holds also for the region where $%
\left( 1-\alpha \right) p\left( \mathbf{z}|\theta \right) /p\left( \mathbf{z}%
|\theta ^{\prime }\right) >1$.

In a companion paper~\cite{nespapericml}, we measured 
the usefulness of importance mixing, and found that it reduces the
number of required fitness evaluations by a factor 5. Additionally,
it reduced the algorithm's sensitivity to the population size.

The computational complexity of importance mixing can be analyzed as
follows. For each generated individual $\mathbf{z}$, the probability $%
p\left( \mathbf{z}|\theta \right) $ and $p\left( \mathbf{z}|\theta ^{\prime
}\right) $ need to be evaluated, requiring $O\left( d^{2}\right) $
computations. The total number of individuals generated is bounded by $%
n/\alpha $ in the worst case, and is close to $n$ on average.

\section{Fitness Shaping}
For problems with wildly fluctuating fitnesses, the gradient
is disproportionately distorted by extreme fitness values, which can lead to
premature convergence or numerical instability. To overcome this problem, we
use \emph{fitness shaping}, an order-preserving nonlinear fitness
transformation function~\cite{nespaper}. The choice of (monotonically increasing)
fitness shaping function is arbitrary, and should therefore be considered to
be one of the tuning parameters of the algorithm. We have empirically found
that ranking-based shaping functions work best for various problems. The
shaping function used for all experiments in this paper was fixed to $%
f^{\prime }(\mathbf{z})=2i-1$ for $i>0.5$ and $f^{\prime }(\mathbf{z})=0$
for $i<0.5$, where $i$ denotes the relative rank of $f\left( \mathbf{z}%
\right) $ in the population, scaled between $0\ldots 1$.

\section{Efficient NES}
Integrating all the algorithm elements introduced above, the Efficient
Natural Evolution Strategy (with block fitness baselines) can be summarized
as%
\begin{equation*}
\begin{tabular}{ll}
\hline
1 & initialize $\mathbf{A}\leftarrow\mathbf{I}$ \\
2 & \textbf{repeat} \\ 
3 & \tab
compute $\mathbf{A}^{-}$, and $\mathbf{C}^{-}=\mathbf{A}^{-}\mathbf{A}%
^{-\top }$ \\ 
4 & \tab
update population using importance mixing \\ 
5 & \tab
evaluate $f\left( \mathbf{z}_{i}\right) $ for new $\mathbf{z}_{i}$ \\ 
6 & \tab
compute rank-based fitness shaping $\hat{f}$\\ 
7 & \tab
\textbf{for} $k=d$ \textbf{to} $0$ \\ 
8 & \tab\tab
compute the exact FIM inverse $\mathbf{F}_{k}^{-}$ \\ 
9 & \tab\tab
$\mathbf{u\leftarrow 0}$, $\mathbf{v}\leftarrow \mathbf{0}$, $%
s_{1}\leftarrow 0$, $s_{2}\leftarrow 0$ \\ 
10 & \tab\tab
\textbf{for} $i=1$ \textbf{to} $n$ \\ 
11 & \tab\tab\tab
$\mathbf{q}\leftarrow \mathbf{F}_{k}^{-}\mathbf{g}^{k}\left( \mathbf{z}%
_{i}\right) $ \\ 
12 & \tab\tab\tab
$\mathbf{u}\leftarrow \mathbf{u}+\hat{f}\left( \mathbf{z}_{i}\right) \mathbf{q}$\\
13 & \tab\tab\tab
$\mathbf{v}\leftarrow \mathbf{v}+\mathbf{q}$ \\ 
14 & \tab\tab\tab
$s_{1}\leftarrow s_{1}+\hat{f}\left( \mathbf{z}_{i}\right) \mathbf{q}^{\top }\mathbf{q}$ \\
15 & \tab\tab\tab
$s_{2}\leftarrow s_{2}+\mathbf{q}^{\top }\mathbf{q}$ \\ 
16 & \tab\tab
\textbf{end} \\ 
17 & \tab\tab
$b^{k}\leftarrow s_{1}/s_{2}$ \\ 
18 & \tab\tab
$\delta \theta ^{k}\leftarrow \mathbf{u}-b^{k}\mathbf{v}$ \\ 
19 & \tab
\textbf{end} \\ 
20 & \tab
$\theta \leftarrow \theta +\eta \delta \theta $ \\ 
21 & \textbf{until} stopping criteria is met \\ \hline
\end{tabular}%
\ \ 
\end{equation*}%
Note that vectors $\mathbf{u}$ and $\mathbf{v}$ in line 18 correspond to $%
\mathbf{F}_{k}^{-}\mathbf{G}^{k}\mathbf{f}$ and $\mathbf{F}_{k}^{-}\mathbf{G}%
^{k}\mathbf{1}$, respectively. Summing up the analysis from previous
sections, the time complexity of processing a single generation is $O\left(
nd^{3}\right) $, while the space complexity is just $O\left( d^{2}+nd\right) 
$, where $O\left( nd\right) $ comes from the need of storing the population.
Assuming that $n$ scales linearly with $d$, our algorithms scales linearly
in space and quadratically in time w.r.t.~the number of parameters, which is 
$O\left( d^{2}\right) $. This is a significant improvement over the original
NES, whose complexity is $O\left( d^{4}\right) $ in space and $O\left(
d^{6}\right) $ in time.

Implementations of eNES are available in both Python and Matlab\footnote{The Python code is part of the PyBrain machine learning library ({\ttfamily www.pybrain.org}) and the Matlab code is available at {\ttfamily www.idsia.ch/\~{}sun/enes.html}}.

\section{Experiments}
\begin{figure}[!ht]
\label{fig:unimodal}
\centerline{ \includegraphics[scale=0.46]{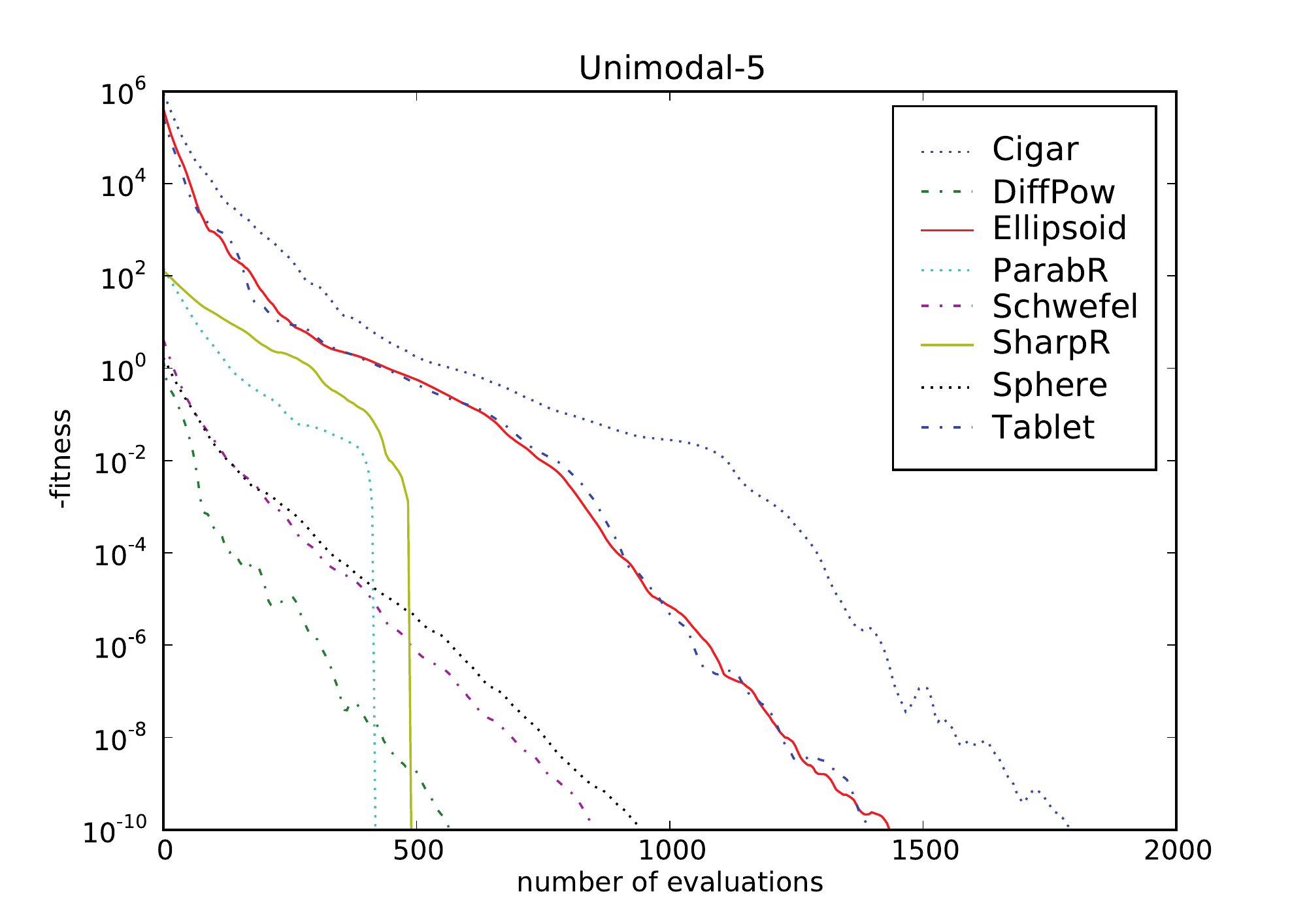}}
\centerline{ \includegraphics[scale=0.46]{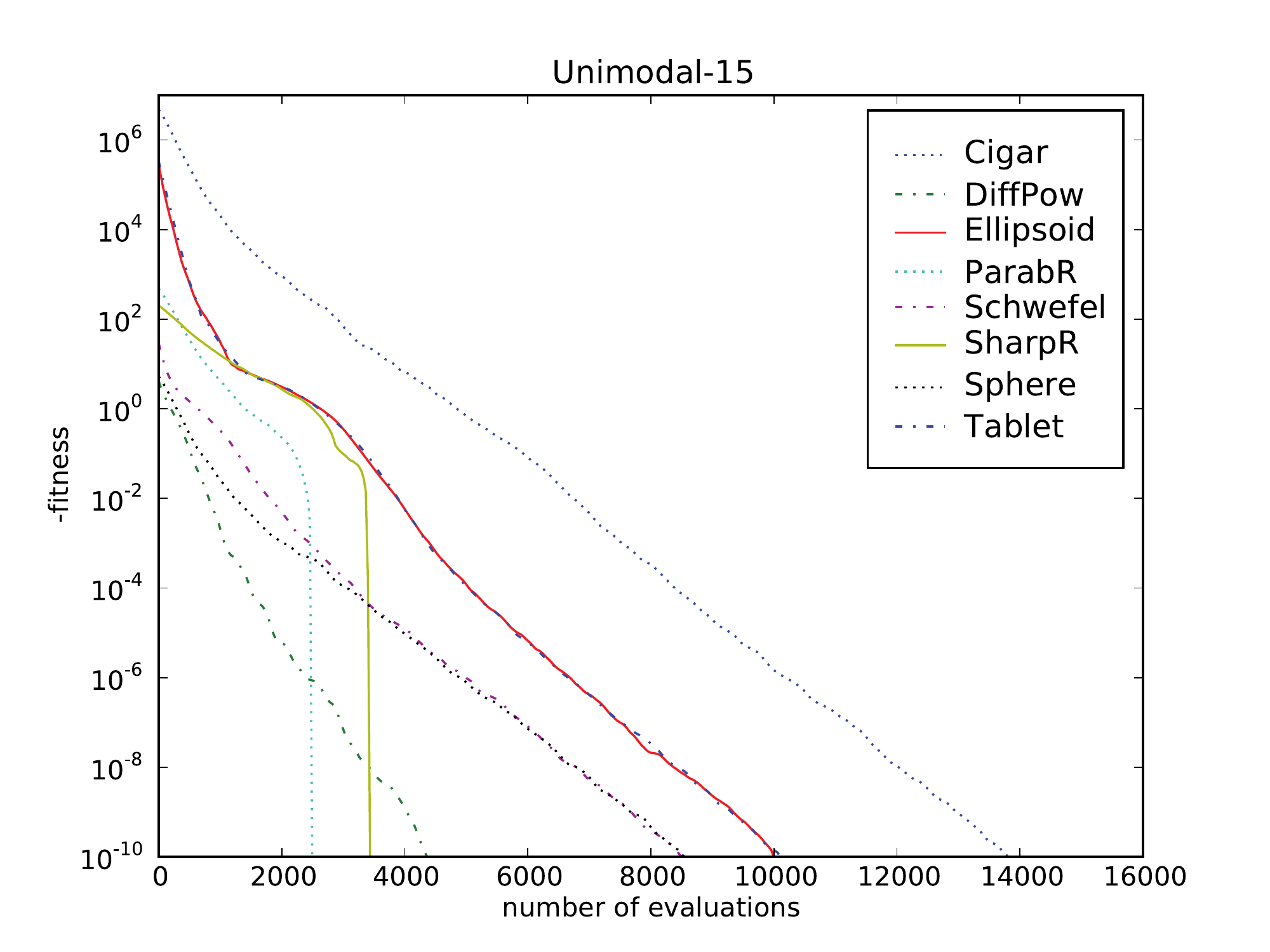}}
\centerline{ \includegraphics[scale=0.46]{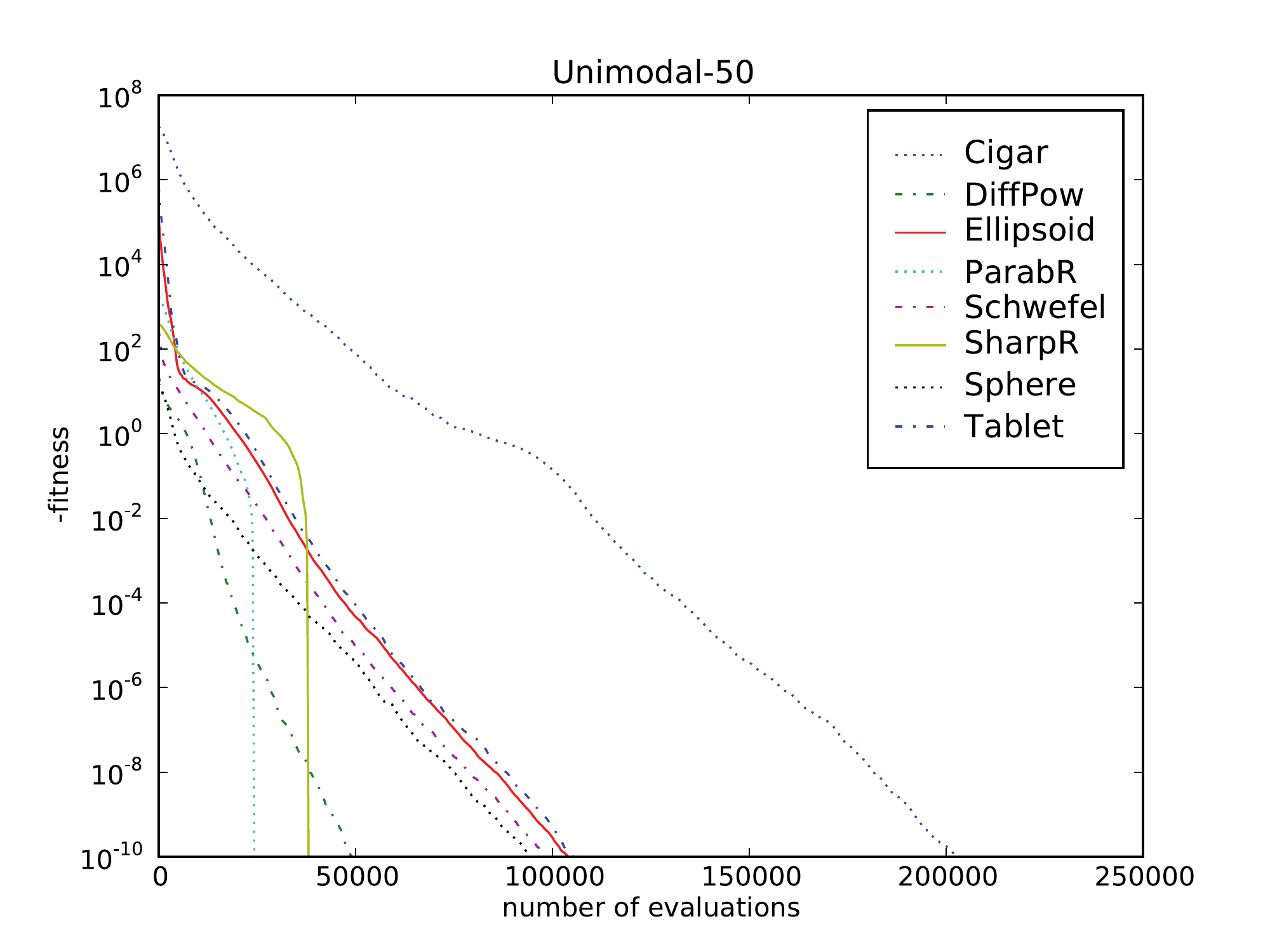}}
\caption{Results on the unimodal benchmark functions for dimension 5, 15 and 50 (from top to bottom).}
\end{figure}

The tunable parameters of Efficient Natural Evolution Strategies are comprised of the population size $n$, 
the learning rate $\eta$, the refresh rate $\alpha$ and the fitness shaping function. 
In addition, three kinds of fitness baselines can be used.

We empirically find a good and robust choice for the learning rate $\eta$ to be $1.0$. 
On some (but not all) benchmarks the performance can be further improved by more aggressive updates.
Therefore, the only parameter that needs tuning in practice is the population
size, which is dependent on both the expected ruggedness of the fitness
landscape and the problem dimensionality. 

\subsection{Benchmark Functions}

We empirically validate our algorithm on 9 unimodal and 4 multimodal
functions out of the set of standard benchmark functions from~\cite{benchmarkset} 
and~\cite{hansenCMA}, that are typically used in the
literature, for comparison purposes and for competitions.
We randomly choose the inital guess at average distance 1 from the optimum.
In order to prevent potentially biased results, we follow~\cite{benchmarkset}
and consistently transform (by a combined rotation and translation) the
functions' inputs, making the variables non-separable and avoiding trivial
optima (e.g.~at the origin). This immediately renders many other
methods virtually useless, since they cannot cope with correlated mutation
directions. eNES, however, is invariant under translation and
rotation. In addition, the rank-based fitness shaping makes it 
invariant under order-preserving transformations of the fitness function.

\subsection{Performance on Benchmark Functions}

We ran eNES on the set of unimodal benchmark functions with
dimensions 5, 15 and 50 with population sizes 50, 250 and 1000, respectively, using $\eta=1.0$ and a
target precision of $10^{-10}$. Figure~1 shows the average performance over
20 runs (5 runs for dimension 50) for each benchmark function. 
We left out the Rosenbrock function on which eNES is one order of magnitude 
slower than on the other functions (e.g. 150,000 evaluations on dimension 15). 
Presumably this is due to the fact that the principal mutation direction is updated too slowly on complex curvatures.
Note that SharpR and ParabR are unbounded functions, which explains the abrupt drop-off.

\begin{figure}[tb]
\label{fig:multimodal}
\centerline{ \includegraphics[scale=0.37]{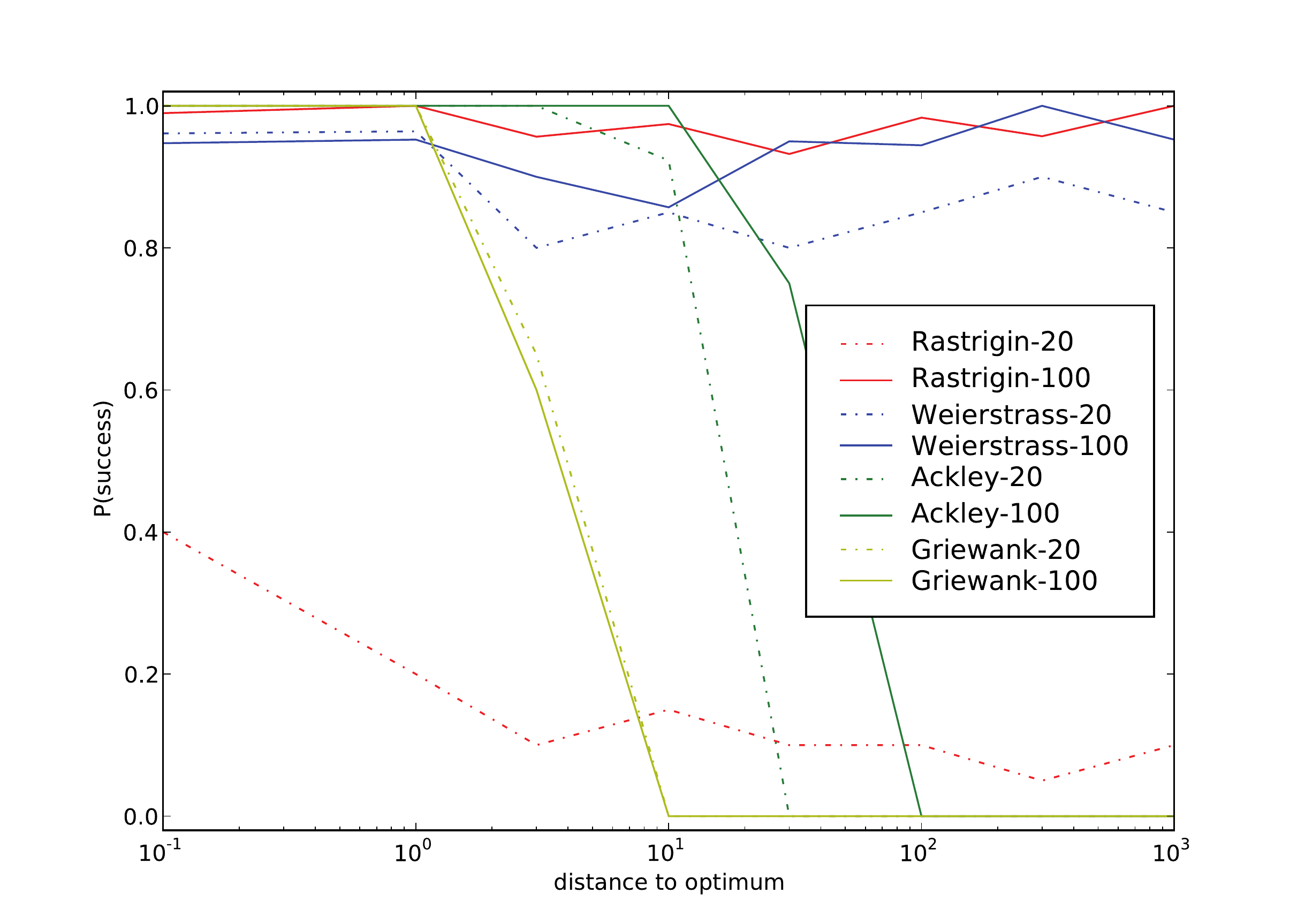}}
\caption{Success percentages varying with initial distances for the
multimodal test functions using population sizes 20 and 100. }
\end{figure}

For the experiments on the multimodal benchmark functions we varied the
distance of the initial guess to the optimum between 0.1 and 1000. Those
runs were performed on dimension 2 with a target precision of $0.01$, since
here the focus was on avoiding local maxima. We compare the results for
population size 20 and 100 (with $\eta=1.0$). Figure~2 shows, for all tested multimodal functions,
the percentage of 100 runs where eNES found the global optimum (as
opposed to it getting stuck in a local extremum) conditioned on the distance
from the initial guess to the optimum.

\begin{figure}[tb]
\label{fig:rastrigin}
\centerline{ \includegraphics[scale=0.5]{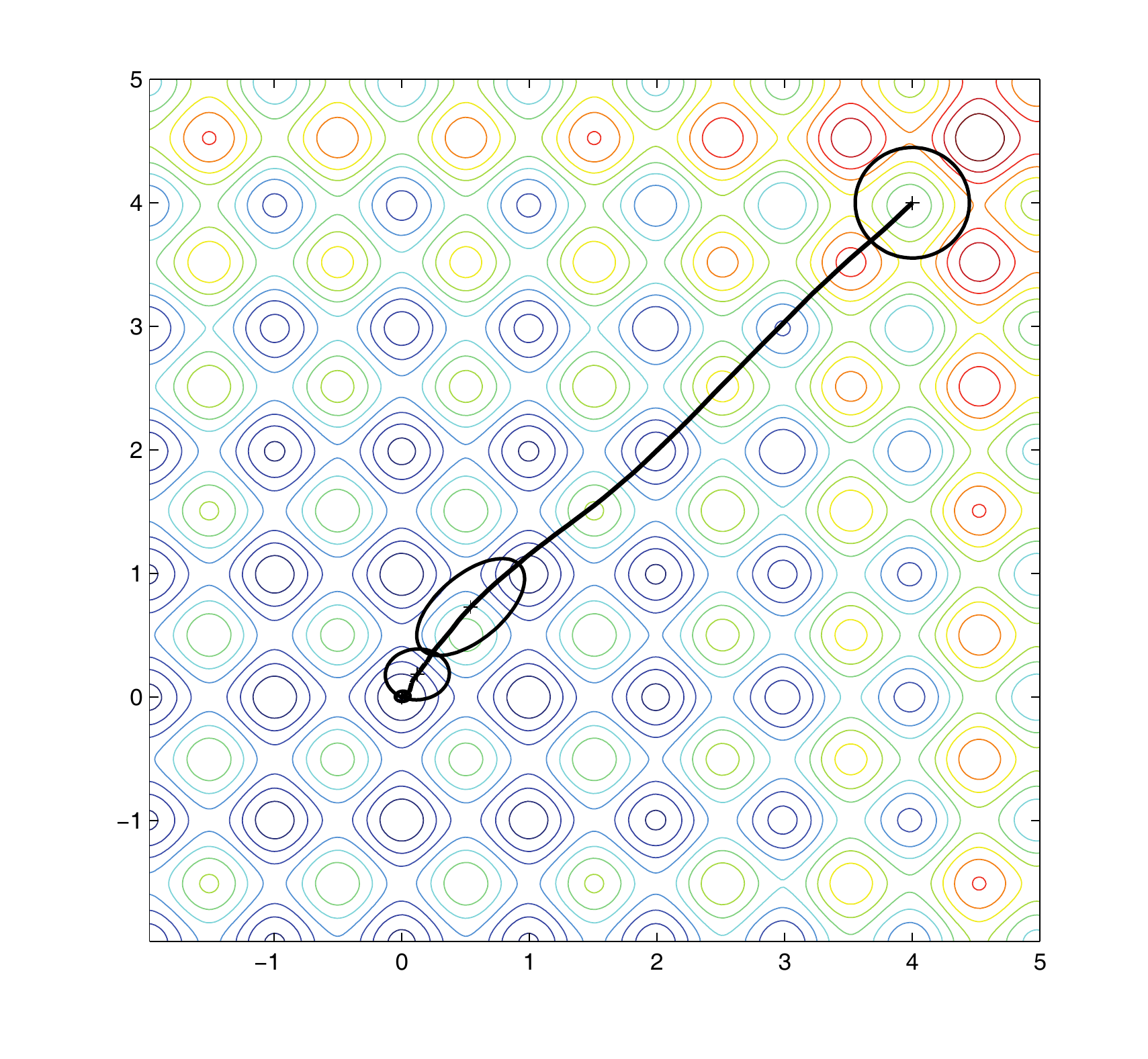}}
\caption{Evolution path and mutation distributions for a typical run on Rastrigin. Ellipsoids correspond to 0.5 standard deviations of the mutation distributions in generations 1, 20, 40.}
\end{figure}

Note that for Ackley and Griewank the
success probability drops off sharply at a certain distance. For Ackley this is 
due to the fitness landscapes providing very little global structure to exploit, 
whereas for Giewank the reason is that the local optima are extremely large, 
which makes them virtually impossible to escape from\footnote{A solution 
to this would be to start with a significantly larger initial $\mathbf{C}$, instead of $\mathbf{I}$}. 
Figure~3 shows the evolution path of a typical run on Rastrigin, and the 
ellipses corresponding to the mutation distribution at different generations, illustrating how eNES 
jumps over local optima to reach the global optimum.

For three functions we find that eNES finds the global optimum reliably, even with
a population size as small as 20. For the other one, Rastrigin, the global optimum is
only reliably found when using a population size of 100.

\section{Discussion}
Unlike most evolutionary algorithms, eNES boasts a relatively clean derivation from first
principles. Using a full multinormal mutation distribution and fitness
shaping, the eNES algorithm is invariant under translation and rotation and under
order-preserving transformations of the fitness function.

Comparing our empirical results to CMA-ES~\cite{hansenCMA}, considered by
many to be the `industry standard' of evolutionary computation, we find that eNES
 is competitive but slower, especially on higher dimensions.
However, eNES is faster on DiffPow for all dimensions.
On multimodal benchmarks eNES is competitive with CMA-ES as well, as compared to the results in~\cite{nespaper}. 
Our results collectively show that eNES can compete 
with state of the art evolutionary algorithms on standard benchmarks. 

Future work will also address the problems of automatically determining good population sizes
and dynamically adapting the learning rate. Moreover, we plan to investigate the possibility 
of combining our algorithm with other methods (e.g.~Estimation of Distribution Algorithms) 
to accelerate the adaptation of covariance matrices, 
improving performance on fitness landscapes where 
directions of ridges and valleys change abruptly (e.g. the Rosenbrock benchmark).

\section{Conclusion}

Efficient NES is a novel alternative to 
conventional evolutionary algorithms, using a natural evolution gradient
to adapt the mutation distribution. 
Unlike previous natural
gradient methods, eNES {\em quickly} calculates 
the inverse of the \emph{exact} Fisher information matrix.
This increases robustness and accuracy of the evolution 
gradient estimation, even in higher-dimensional search spaces. 
Importance mixing prevents unnecessary redundancy
embodied by individuals from earlier generations. 
eNES constitutes a competitive, theoretically well-founded 
and relatively simple method for artificial evolution. 
Good results on standard benchmarks 
affirm the promise of this research direction.

\section{Acknowledgments}
This research was funded by SNF grants 200020-116674/1, 200021-111968/1 and 200021-113364/1.

%
%
\end{document}